# A Comparative Study on Feature Selection for a Risk Prediction Model for Colorectal Cancer

Nahúm Cueto-López[a], Maria Teresa García-Ordás[a], Veronica Dávila-Batista[b], Víctor Moreno[c], Nuria Aragonés[d], Rocío Alaiz-Rodríguez[a,*]

[a]*Department of Electrical, Systems and Automatic Engineering, Universidad of León. Campus de Vegazana s/n, 24071 León*
[b] *Centro de Investigación Biomédica en Red (CIBER). Grupo Investigación Interacciones Gen-Ambiente y Salud (GIIGAS), Spain*
[c]*Unit of Biomarkers and Susceptibility, Cancer Prevention and Control Programme, Catalan Institute of Oncology-IDIBELL, L'Hospitalet de Llobregat, Spain.Department of Clinical Sciences, Faculty of Medicine, University of Barcelona, Barcelona, Spain.CIBER Epidemiologa y Salud Pblica (CIBERESP), Madrid, Spain.*
[d]*Epidemiology Section, Public Health Division, Department of Health of Madrid. CIBER Epidemiología y Salud Pública (CIBERESP), Madrid, Spain;*

**Abstract**

*Background and objective*: Risk prediction models aim at identifying people at higher risk of developing a target disease. Feature selection is particularly important to improve the prediction model performance avoiding overfitting and to identify the leading cancer risk (and protective) factors. Assessing the stability of feature selection/ranking algorithms becomes an important issue when the aim is to analyze the features with more prediction power.
*Methods*: This work is focused on colorectal cancer, assessing several feature ranking algorithms in terms of performance for a set of risk prediction models (Neural Networks, Support Vector Machines (SVM), Logistic Regression, k-Nearest Neighbors and Boosted Trees). Additionally, their robustness is evaluated following a conventional approach with scalar stability metrics and a visual approach proposed in this work to study both similarity among fea-ture ranking techniques as well as their individual stability. A comparative analysis is carried out between the most relevant features found out in this study and features provided by the experts according to the state-of-the-art knowledge.

[*]Corresponding Author
 *Email addresses:* ncuetl00@unileon.es (Nahúm Cueto-López), mgaro@unileon.es (Maria Teresa García-Ordás), vdavb@unileon.es (Verónica Dávila-Batista), v.moreno@iconcologia.net (Víctor Moreno), nuria.aragones@salud.madrid.org (Nuria Aragonés), rocio.alaiz@unileon.es (Roc´ıo Alaiz-Rodr´ıguez )

*Results*: The two best performance results in terms of Area Under the ROC Curve (AUC) are achieved with a SVM classifier using the top-41 features selected by the SVM wrapper approach (AUC=0.693) and Logistic Regression with the top-40 features selected by the Pearson (AUC=0.689). Experiments showed that performing feature selection contributes to classification performance with a 3.9% and 1.9% improvement in AUC for the SVM and Logistic Regression classifier, respectively, with respect to the results using the full feature set. The visual approach proposed in this work allows to see that the Neural Network-based wrapper ranking is the most unstable while the Random Forest is the most stable.

*Conclusions*: This study demonstrates that stability and model performance should be studied jointly as Random Forest turned out to be the most stable algorithm but outperformed by others in terms of model performance while SVM wrapper and the Pearson correlation coefficient are moderately stable while achieving good model performance.

## 1. Introduction

ColoRectal Cancer (CRC) is ranked third and second among all cancer incidences in men and women, respectively worldwide [13]. It is the fourth leading cause of cancer death in the world, accounting for over one million new cases of colorectal cancer (CRC) diagnosed and more than 880000 deaths in 2018 [14]. Globally, CRC has increased steadily worldwide since the 1960s but there is substantial geographical variation in incidence and mortality rates across the world. The distribution of CRC varies widely, with more than two-thirds of all cases and about 60% of all deaths occurring in countries with a high human development index [15]. Thus, CRC rates are rising in countries that are undergoing rapid economic development [8] due to economic transitions and the relation with lifestyle issues such as diet, physical inactivity and obesity [20]. On the other hand, preventive screening and specialized care are changing the trends in reported mortality in other countries [2]. Risk prediction models becomes an important tool to iden- tify people at increased risk of developing CRC and uncover the risk factors (features, in the context of this work) for this disease [23].

Feature selection is a key step in many classification problems [17, 36, 4].The size of the training data set needed to calibrate a model grows expo- nentially with the number of dimensions but the number of instances may be limited due to the cost of data collection. In particular, in cancer risk pre- diction applications, reducing the data dimensionality can avoid overfitting and improve model performance [3, 10, 27, 11]. Additionally, the process of knowledge discovery from the data is simplified when the unwanted noisy and irrelevant features are removed.

Numerous works have examined feature selection with respect to classi- fication performance [31, 9, 25], but a problem that arises in many practical problems is that small variations in the data lead to different outcomes of the feature selection algorithm. Perhaps the disparity among different re- search findings has made the study of the robustness (or stability) of feature selection techniques a topic of recent interest [37, 18, 39, 28, 1].

When developing a cancer risk prediction model, performance is not the only goal but also extracting a feature subset of the most relevant features in order to better understand the data and the underlying process. Thus, in this work, we assess several feature ranking techniques in the context of a colorectal cancer prediction model. Fig. 1 shows the proposed system diagram. From the whole dataset, different data samples are extracted. The feature ranking technique applied on each of these subsets lead to different feature rankings. The feature selection method is evaluated both with respect to classifier performance (after combining these individual feature rankings) and also with respect to their stability (or robustness), trying to measure the similarity among the rankings.

Several (scalar) metrics [22, 19] have been proposed to evaluate the sta- bility of the feature selection process. In this work, assessment is conducted following some of these metrics and we also propose a graphical approach [5] that enables us to analyze the similarity between feature ranking tech- niques as well as their individual stability. We also compare the performance achieved with the risk prediction models that use features selected with fea- ture selection techniques and those models that rely on features that the experts consider to be state-of-the-art.

The rest of this paper is organized as follows: Section 2 describes the feature selection process and its stability. Experimental evaluation is shown

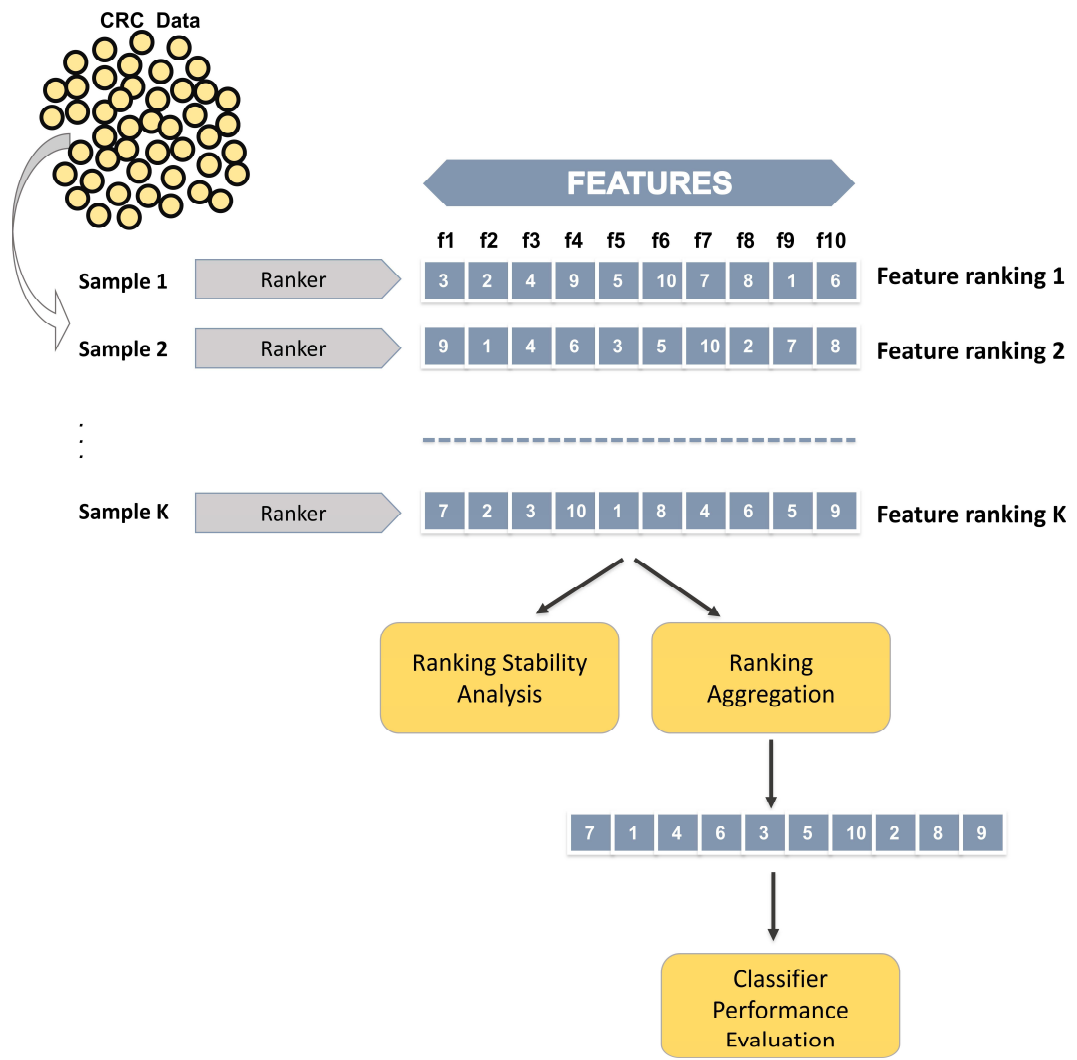

Figure 1: The proposed system diagram for the assessment of feature ranking techniques.

in Section 3 and discussion in Section 4. Finally, Section 5 summarizes the main conclusions.

## 2. Methods

Feature selection techniques measure the importance of a feature or a set of features according to a given measure. There are many goals of these techniques, but the most important ones are [36]: (a) to mitigate the curse of dimensionality, (b) to gain a deeper insight into the underlying processes that generated the data and (c) to provide faster and more cost-effective prediction models.

Consider a training dataset $D = \{(\mathbf{x}_i, d_i), i = 1, \ldots, M\}$ with M examples and a class label $d$ associated with each sample. Each instance $\mathbf{x}_i$ is a $p$-dimensional vector $\mathbf{x}_i = (x_{i1}, x_{i2}, \ldots x_{ip})$ where each component $x_{ij}$ represents the value of a given feature $f_j$ for that example $i$.

Consider now a feature ranking algorithm that leads to a ranking vector $\mathbf{r}$ with components

$$\mathbf{r} = (r_1, r_2, r_3, \ldots, r_p) \tag{1}$$

where $1 \leq r_i \leq p$. Note that 1 is considered the highest rank.

Consider also a top-k list as the outcome of a feature selection technique

$$\mathbf{s} = (s_1, s_2, s_3, \ldots, s_p), s_i \in \{0, 1\} \tag{2}$$

where 1 indicates the presence of a feature and 0 the absence and $\sum_{i=1}^{p} s_i = k$.

Feature selection techniques usually generate a full ranking of features. These rankings, however, can be converted in top-k lists that contain the most important k features. Converting a ranking output into a feature subset is easily conducted according to

$$s_i = \begin{cases} 1 & \text{if } r_i \leq k \\ 0 & \text{if } otherwise \end{cases}$$

In the context of classification, the feature selection or ranking techniques can be basically organized into three categories [17, 6]: *filter*, *wrapper* and *embedded* approaches. The *filter* methods rely on general characteristics of the training data to rank the features according to some metric without involving any learning algorithm. The *wrapper* approaches incorporate the interaction between the feature selection process and the classification model, in order to determine the value of a given feature subset. Finally, in the *embedded* techniques, the feature search mechanism is built into the classifier model and are therefore specific to a given learning algorithm. The ranking methods studied in this work are briefly described next.

*2.1. Feature Selection with filters*

Within this category, we consider the well-known Relief algorithm and the simple Pearson correlation coefficient that has proven to be very effective, even though it does not remove feature redundancy [16].

*Relief.*
The basic idea of the Relief algorithm is to reweigh features according to their ability to distinguish examples of the same and different classes that are near to each other [38].

*Pearson correlation coefficient.*
This method looks at how well correlated each feature is with the class target. If a feature is highly correlated with one of the classes, then we can assume that it is useful for classification purposes [38].

*2.2. Feature Selection with Wrapper approaches*

Wrapper methods use the performance of a learning algorithm to assess the usefulness of a feature set. Either they iteratively discard features with the least discriminant power or they add the best features according to model performance [17]. However, wrapper approaches are more computationally intensive than filter methods.

In this work, we evaluate two wrapper approaches that measure importance of a feature set based on the performance of a Support Vector Machine an a Neural Network with a MultiLayer Perceptron architecture. In both cases, model performance is estimated by the Area under the ROC (Receiver Operating Characteristic) curve (AUC), where the ROC curve plots the true positive rate against the false positive rate.

*2.3. Feature Selection with Embedded approaches*

In this work we evaluate two embedded feature ranking strategies based on Random Forests and SVM, respectively.

*SVM with Recursive Feature Elimination (SVM-RFE).*
SVM-RFE [17] determines which features provide the best contribution to the precision of the model while it is being created. This increases the performance in terms of time compared with the wrapper techniques.

*Random Forests (RF).*
Every node of the multiple decision trees that conforms the RF is a condition over a single feature. Taking into account the performance of the nodes, a ranking of features can be easily created [17].

*2.4. Stability of Feature Selectors*

An important property of a feature selection method is its stability [18, 26, 34]. The fact that under small variations in the supplied data, the outcome of the feature selection technique varies (either a full ranked list or a top-k list), makes the conclusions derived from it unreliable.

Consider we run a feature ranking algorithm $K$ times. Results can be gathered in a matrix $\mathbf{A}$ with elements $r_{ij}$ with $i = 1, \ldots, p$ and $j = 1, \ldots, K$ that indicate the rank assigned in the run-$j$ for feature-$i$. The same applies to a feature selector.

In general, stability is quantified as follows: Given a set of rankings (subsets), pairwise similarities are computed and then, reduced to a single metric by averaging. These (scalar) metrics can be seen as projections to one dimensional space and its use only shows where the feature selector stands in relation to the stable and the random ranking algorithm. In this paper, we also want to illustrate and motivate the use of graphical methods as a simple alternative approach to evaluate the stability of feature ranking algorithms. We will show how the projection to two dimensions allow the evaluation of the similarity between feature ranking algorithms as well as their stability.

Next, we discuss different approaches to quantify the robustness of feature selector or ranking algorithms by (1) a conventional analysis and (2) a visual-based study.

*2.4.1. Conventional Stability Analysis*

To study the stability of the feature ranking or selection techniques several metrics have been proposed.

*Similarity Measures.*
Consider $\mathbf{r}$ and $\mathbf{r}^t$ the output of a feature ranking technique applied to two subsamples of $\mathcal{D}$. The most widely used metric to measure the similarity between two ranking lists is the Spearman's rank correlation coefficient (SR) [30]. The SR between two ranked lists $\mathbf{r}$ and $\mathbf{r}^t$ is defined by

$$SR(\mathbf{r}, \mathbf{r}^t) = 1 - 6 \sum_{i=1}^{p} \frac{(r_i - r_i^t)^2}{p(p^2 - 1)} \quad (3)$$

where $r_i$ is the rank of feature-$i$. $SR$ values range from 1 to 1. It takes the value one when the rankings are identical and the value zero when there is no correlation.

When we attempt to measure the distance between two top-k lists **s** and $\mathbf{s}^t$ with the most relevant k features, several metrics have been presented (for details see [30]). In this work we use the Jaccard stability index (JI) that can be defined as

$$JI(\mathbf{s}, \mathbf{s}^t) = \frac{|\mathbf{s} \wedge \mathbf{s}^t|}{|\mathbf{s} \vee \mathbf{s}^t|} = \frac{r}{l} \quad (4)$$

where **s** and $\mathbf{s}^t$ are the two feature subsets, $r$ is the number of features that are common in both lists and $l$ the number of features that appear only in one of the two lists. The JI lies in the range $(0, 1)$.

*The Stability for a Set of Rankings or Lists.*

When it comes to evaluate the stability of a feature selection (or ranking) algorithm that provides several results $A = \{\mathbf{r}_1, \mathbf{r}_2, \ldots \mathbf{r}_K\}$, the most popular approach is to compute pairwise similarities and average the results, what leads to a single scalar value.

$$S(A) = \frac{2}{K(K-1)} \sum_{i=1}^{K-1} \sum_{j=i+1}^{K} S_M(r_i, r_j) \quad (5)$$

where $S_M$ may be any distance metric like the Spearman rank correlation coefficient, Jaccard stability index [22, 30] or Kuncheva's stability index [24], for example.

*2.4.2. Visual based Stability Analysis*

The outcome of a feature ranking algorithm can be interpreted as a point in a high dimensional space (with $p$ dimensions). The stability of a ranking feature selector is commonly measured as the dissimilarity or distance between different outcomes of the same feature selector on slightly different datasets. As mentioned above, stability is assessed computing pairwise similarities between points in that high dimensional space and averaging the results. In this case, the ranking data is turned into a single number (projected to one dimension) and the algorithms are compared on the basis of this scalar metric. This only allows the comparison of the feature selector with respect to a reference: the random ranking and the completely stable ranking.

Note that if we change from a projection, to a space with one dimension, into a space with two or more dimensions, we have a visual representation that allows to establish comparisons with respect to the random selector as well as comparisons of each feature selector to the others.

In order to study the stability with a visual-based approach, different alternatives could be used, depending on the amount of information available. Note that, even simple visualization approaches like histograms or scatter graphs allow the depiction of the results in a convenient way to ease result interpretation. They have some limitations as the number of dimensions increases. In this case, a dimensionality reduction technique like MultiDimensional Scaling (MDS) [12], that preserves as mucho of the original data structure as possible, seems more convenient. It allows the projecetion of data from a high dimensional space to a 2D or 3D space while preserving the distance in the original high dimensional space.

## 3. Experimental Results with a Colorectal Cancer Dataset

In this section, we build a CRC risk prediction model assessing several feature ranking algorithms. The evaluation is conducted in terms of the classification performance (predictive power) and the robustness of the ranking algorithms.

### 3.1. Colorectal Cancer Dataset

Experimental results were carried out with a CRC dataset from the MCC-Spain study [21]. MCC-Spain is a multicentric case-control study with population controls aiming to evaluate the influence of environmental exposures and their interaction with genetic factors in common tumors in Spain (prostate, breast, colorectal, gastroesophageal and chronic lymphocytic leukemia).All participants signed an informed consent. Approval for the study was obtained from the ethical review boards of all recruiting centers.[7]. Instances with missing values have been removed leading to a dataset with 3295 instances: 2230 are controls, while the other 1065 are cases. Each individual is described by 100 features: 47 genetic variables (Single Nucleotide Polymorphisms -SNPs), 48 environmental factors including red meat, vegetable consumption, BMI, physical activity, alcohol consumption and 5 variables regarding family history of CRC, sex, age, level of education and race.

Next, the variables considered in this study are listed.

- **SNP**: rs 10411210, rs 10505477, rs 1057910, rs 10761659, rs 10795668, rs 10883365, rs 10936599, rs 11169552, rs 11209026, rs 12035082, rs 13361189, rs 16940372, rs 17309827, rs 1800588, rs 1801282, rs 2470890, rs 2542151, rs 268, rs 328, rs 3802842, rs 3824999, rs 405509, rs 439401, rs 4444235, rs 4774302, rs 4775053, rs 4779584, rs 4925386, rs 4939827, rs 5275, rs 5771069, rs 5934683, rs 6083, rs 6687758, rs 6691170, rs 6887695, rs 6983267, rs 7014346, rs 7136702, rs 7259620, rs 744166, rs 762551, rs 7758229, rs 916977, rs 961253, rs 9858542, rs 9929218.

- **Environmental factors**. physical activity, BMI, alcohol consumption, smoking. Dietary factors: consumption of vegetable, red meat, legume, fruit, cereals, fish, dairy, oil, calcium, carotenoids, cholesterol, edible, total energy, ethanol in the past decade, ethanol in the present, monounsaturated fats, polyunsaturated fats, saturated fats, total fats, folic acid, glucids, total intake in grams, Iron, magnesium, niacin, phosphorus, potassium, fiber, animal protein, vegetable protein, total protein, retinoids, sodium, digestible sugars, polysaccharides, vit A, vit B1, vit B12, vit B2, vit B6, vit C, vit D, vit E, water and zinc.

- **Other factors**: family history of CRC, sex, age, level of education, race.

*3.2. Assessment of the predictive power*

We have chosen several classifiers to be calibrated as CRC prediction models, all of them very different from each other: Logistic Regression, k-Nearest Neighbors, Neural Networks with a Multilayer Perceptron architecture, Support Vector Machines and Boosted Trees.

*Logistic Regression* (LR). We trained a logistic regression classifier using a binomial distribution of the response variable.

*k-Nearest Neighbors* (k-NN).
Nearest neighbors with k=47 are extracted using the cosine distance (one minus the cosine of the included angle between observations). Features are normalized with zero mean and standard deviation equal to one.

*Support Vector Machines* (SVM). We have used a SVM with a gaussian kernel and training was performed using the Sequential Minimal Optimiza-tion routine. A standardization of the training and test set was carried out.

*Boosted Trees* (BT). The AdaBoostM1 ensemble aggregation method is used with a learn rate of 0.1. In our case, we have fixed the maximal number of decision splits (or branch nodes) per tree to 20.

*Neural Networks* (NN). We evaluate a three layer neural network with a logistic sigmoid activation function for the hidden and the output layers. The network has been trained using the scaled conjugate gradient backpropagation algorithm and the cross entropy is the cost function minimized in the training stage. Several combinations of neurons in the hidden layer and different number of training cycles have been assessed with all the descriptors, in order to find the optimal network configuration (100 training cycles and 4 nodes in the hidden layer).

In order to avoid overfitting, the NN is trained with 60% for training, 20% for validation and the other 20% for testing. The AUC and classification accuracy for the NN classifier is estimated as the average of three runs. For the other classifiers performance is estimated using 5-fold cross validation. Table 1 shows the AUC achieved with the different classifiers with the whole set of features (without performing feature selection), what gives us baseline performance information.

Table 1: AUC for different classifiers with the original dataset (without performing feature selection).

| Number of features | LR | k-NN | NN | SVM | BT |
|---|---|---|---|---|---|
| 100 | 0.676 | 0.624 | 0.540 | 0.667 | 0.671 |

Classifiers were also trained with the most important features selected by different ranking approaches. We have chosen six feature rankers representative of the three main categories. Two of the feature selection algorithms are based on a filter approach (Relief and the Pearson correlation coefficient), another two follow a wrapper approach (SVM and Neural Networks guided by the AUC classifier performance) and two are embedded approaches (SVM-RFE and RF).

In our experimental setting, the feature ranking algorithms are launched with 70% of the data randomly extracted from the whole dataset. Seven runs of this process resulted in a total of seven rankings. The ranking used for this purpose resulted from the aggregation of the 7 rankings generated from the different runs of the algorithm by computing their median value. Feature ranking was carried out with Python for the embedded approaches and with MATLAB for the remaining methods

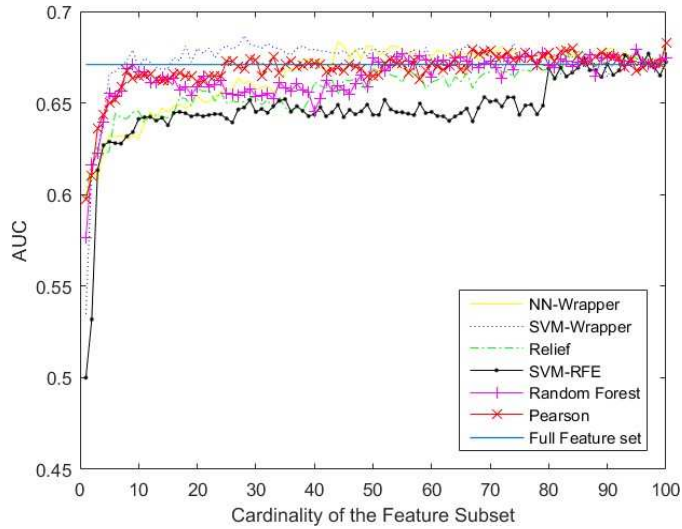

Figure 2: Classifier performance (AUC) with the full feature set and different cardinality of the feature subset for different classifiers: BT.

Fig. 2 to Fig. 6 plot the AUC of five different classifiers trained with a number of features ranging from 1 to 100, selected according to their relevance with the six different ranking algorithms assessed in this work. Additionally, it shows the classifier performance achieved with the full feature set. Except for the NN classifier that is very unstable, it can be seen that performance increases as we increase the number of features used as predictors. Note, however, that the AUC for some classifiers like SVM, k-NN and LR start to degrade from one point onwards as more new features - the most irrelevant or redundant features - are added. Thus, the AUC for k-NN classifier is 0.673 with the top-22 features selected by SVM-wrapper approach and decreases to 0.642 with 100 features (see Fig. 3). Similarly, the AUC achieved by the LR method increases from 0.676 with all the features to 0.689 with the top-40 most relevant features selected with Pearson correlation coefficient (Fig. 4).

In order to simplify the analysis of the techniques that provide the most informative features, Table 2 records, for each classifier, the best feature selection strategy according to the AUC metric. This is carried out for the top-40, top-55 and top-70 features. Thus, Table 2 collects the three best feature sets up to a cardinality of 40 features that lead to the best perfor- mance for each one of the classifiers. Likewise, the best feature sets up to a cardinality of 55 and 70. It can be seen that considering up to 70 features, the SVM classifier work best with the following three sets: top-41 SVM- wrapper, top-45 SVM-wrapper and top-56 SVM-wrapper. It turns out that the SVM-wrapper approach provides the most relevant features in most cases followed by the feature selection technique based on the Pearson correlation coefficient.

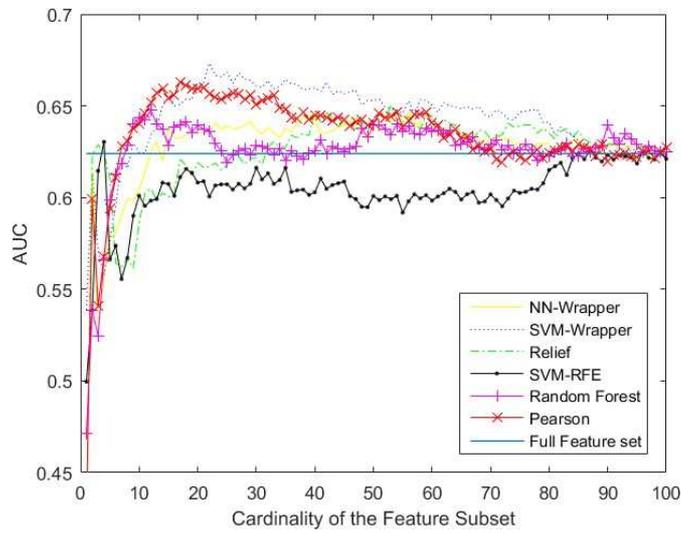

Figure 3: Classifier performance (AUC) with the full feature set and different cardinality of the feature subset for different classifiers: KNN.

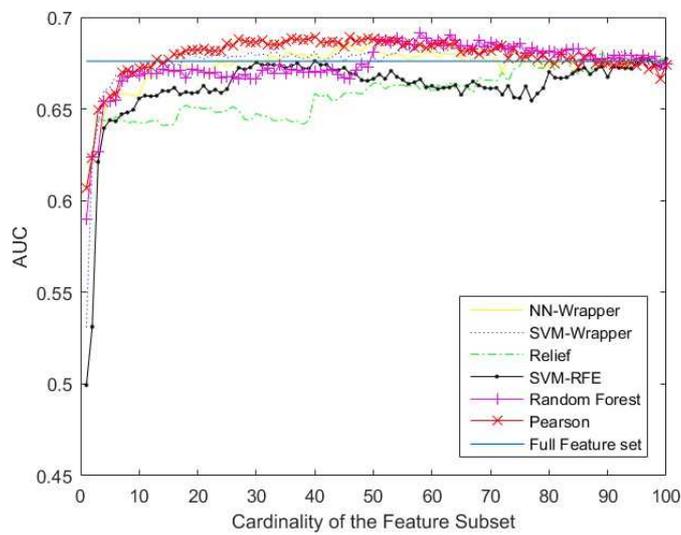

Figure 4: Classifier performance (AUC) with the full feature set and different cardinality of the feature subset for different classifiers: LR.

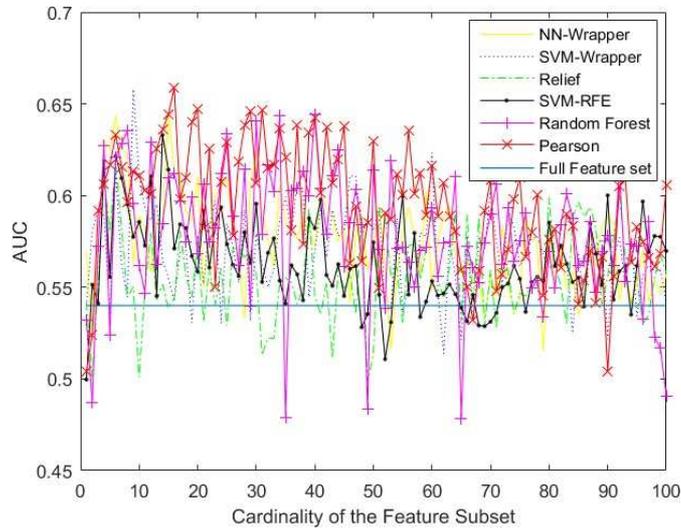

Figure 5: Classifier performance (AUC) with the full feature set and different cardinality of the feature subset for different classifiers: NN.

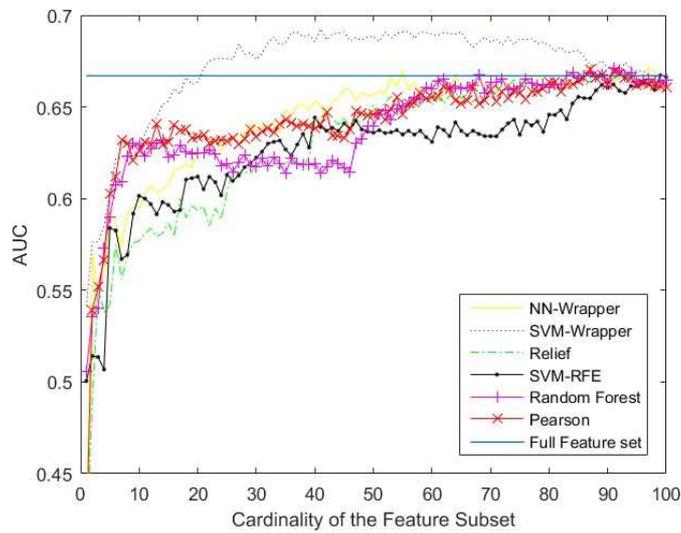

Figure 6: Classifier performance (AUC) with the full feature set and different cardinality of the feature subset for different classifiers: SVM.

Table 2: Best Feature selection techniques for different top-k lists and classifiers.

|  | SVM | | | BT | | | kNN | | |
|---|---|---|---|---|---|---|---|---|---|
|  | #Features | Rank | AUC | #Features | Rank | AUC | #Features | Rank | AUC |
| Top 40 | 36 | SVM-wrapper | 0.691 | **28** | **SVM-wrapper** | 0.687 | **22** | **SVM-wrapper** | 0.673 |
| | **37** | **SVM-wrapper** | 0.691 | 29 | SVM-wrapper | 0.682 | 25 | SVM-wrapper | 0.669 |
| | 39 | SVM-wrapper | 0.689 | 32 | SVM-wrapper | 0.682 | 27 | SVM-wrapper | 0.668 |
| Top 55 | **41** | **SVM-wrapper** | 0.693 | **28** | **SVM-wrapper** | 0.687 | **22** | **SVM-wrapper** | 0.673 |
| | 45 | SVM-wrapper | 0.691 | 32 | SVM-wrapper | 0.682 | 25 | SVM-wrapper | 0.669 |
| | 55 | SVM-wrapper | 0.691 | 44 | NN-wrapper | 0.684 | 27 | SVM-wrapper | 0.668 |
| Top 70 | **41** | **SVM-wrapper** | 0.693 | **28** | **SVM-wrapper** | 0.687 | **22** | **SVM-wrapper** | 0.673 |
| | 45 | SVM-wrapper | 0.691 | 32 | SVM-wrapper | 0.682 | 25 | SVM-wrapper | 0.669 |
| | 56 | SVM-wrapper | 0.691 | 44 | NN-wrapper | 0.684 | 27 | SVM-wrapper | 0.668 |

|  | LR | | | NN | | |
|---|---|---|---|---|---|---|
|  | #Features | Rank | AUC | #Features | Rank | AUC |
| Top 40 | 27 | Pearson | 0.688 | 9 | SVM-wrapper | 0.658 |
| | 36 | Pearson | 0.688 | **16** | **Pearson** | 0.659 |
| | **40** | **Pearson** | 0.689 | 20 | Pearson | 0.647 |
| Top 55 | **40** | **Pearson** | 0.689 | 9 | SVM-wrapper | 0.658 |
| | 46 | Pearson | 0.689 | **16** | **Pearson** | 0.659 |
| | 49 | Pearson | 0.689 | 20 | Pearson | 0.647 |
| Top 70 | 40 | Pearson | 0.689 | 9 | SVM-wrapper | 0.658 |
| | **58** | **RF** | 0.692 | **16** | **Pearson** | 0.659 |
| | 63 | RF | 0.691 | 20 | Pearson | 0.647 |

### 3.3. Ranking Stability Analysis

The stability of six feature ranking algorithms is evaluated in this section. The feature ranking algorithm was launched with 70% of the data randomly extracted from the whole dataset. Seven runs of this process resulted in a total of $K = 7$ rankings.

### 3.3.1. Traditional Stability Analysis

The stability of the feature ranking algorithms can be evaluated with metrics like the Spearman's rank correlation coefficient (SR). In this case, we have computed the $\frac{7(7-1)}{2}$ pairwise similarities for each algorithm to end up averaging these computations according to Eq.(5). The SR is recorded in Table 3 where it can be seen that RF is the most stable (0.712) ranking algorithm, whereas NN-Wrapper is quite unstable (0.036).

Table 3: Stability of a set with 7 full rankings assessed through average pairwise similarities with the Spearman's rank correlation coefficient (SR).

| Pearson | Relief | SVM Wrapper | NN Wrapper | SVM RFE | RF |
|---|---|---|---|---|---|
| 0.251 | 0.280 | 0.078 | 0.036 | 0.240 | 0.712 |

Table 4: Stability of a set with 7 top-k lists assessed through average pairwise similarities with the Jaccard index for different values of $k$.

| $k$ | Pearson | Relief | SVM Wrapper | NN Wrapper | SVM RFE | RF |
|---|---|---|---|---|---|---|
| 10 | 0.711 | 0.312 | 0.406 | 0.179 | 0.317 | 0.550 |
| 20 | 0.783 | 0.434 | 0.440 | 0.171 | 0.354 | 0.695 |
| 30 | 0.745 | 0.565 | 0.515 | 0.207 | 0.400 | 0.804 |
| 35 | 0.767 | 0.633 | 0.528 | 0.242 | 0.424 | 0.831 |
| 40 | 0.784 | 0.705 | 0.519 | 0.278 | 0.466 | 0.867 |
| 50 | 0.733 | 0.733 | 0.549 | 0.364 | 0.567 | 0.981 |
| 60 | 0.715 | 0.729 | 0.599 | 0.446 | 0.664 | 0.844 |
| 70 | 0.746 | 0.744 | 0.642 | 0.542 | 0.773 | 0.828 |
| 80 | 0.788 | 0.792 | 0.73 | 0.666 | 0.881 | 0.882 |
| 90 | 0.847 | 0.875 | 0.833 | 0.818 | 0.888 | 0.963 |
| 100 | 1 | 1 | 1 | 1 | 1 | 1 |
| Average for k from 1 to 100 | 0.776 | 0.665 | 0.588 | 0.431 | 0.597 | 0.826 |

The Jaccard index allows to study the stability of a feature subset that contains the top-$k$ feature lists. Table 4 shows the Jaccard index for the selection of feature subsets with cardinality that varies from 10 to 100 and the average in the last row. The results confirm that the NN-Wrapper method is very unstable and RF is very stable. Looking at stability and classifier performance jointly, results demonstrate that RF was the most stable technique, but it performed worse than other rankers in terms of model performance.

SVM-wrapper and the Pearson correlation coefficient performed moderately in terms of robustness and are the best ranking technique in terms of model performance.

The analysis based on a single metric does not allow, however, to say anything about how similar the rankings provided by the different algorithms are. Typical questions we would like to answer are: (i) Which feature ranking algorithms provide similar rankings?, (ii) Which algorithm is more stable for a certain range of $k$ values?. Analyzing directly the results gathered in Table 4 does not seem straightforward.

*3.3.2. Visual Stability Analysis*

A simple plot helps to see the relative and absolute stability of the feature selectors. Fig. 7 highlights that their relative stability changes with the value of $k$. In general terms, RF and Pearson appears to be the most stable algorithm. Note also that the stability of the SVM-RFE approach for low values of $k$ is very low. No reliable information of the most relevant factors can be extracted from just a single run of the algorithm. It would be desirable to aggregate the rankings in order to get a more representative ranking. Likewise, NN-Wrapper is very unstable.

MDS [12] is used in this section to visualize the feature selectors in a graph so that comparisons between all of them can be established.

All the results gathered in the experiment can be organized as a set of 42 points (6 algorithms x 7 runs each one) defined a 100-dimensional space. These points are projected to a 2D space using MDS. The distance between points is calculated with the Spearman's rank coefficient and the stress criterion is normalized with the sum of squares of the dissimilarities.

After the projection, each outcome of the algorithm is represented by two coordinates (x,y) and the similarities among feature selector can be analyzed in Fig.7(b). Regarding stability, it can be observed that the points that correspond to the NN-Wrapper are very scattered. In other words, this is the most unstable feature selector. The outcomes of RF, however, are clustered together. The same applies to the Pearson feature selector. This figure also allows to see that Pearson generates similar ranking to RF. Note also that SVM-RFE and Releif are very distant to these two methods while SVM-Wrapper falls somewhere in between.

Stability should be studied jointly with the capability of the selected features to predict the target class. This is crucial in order to provide reliable information to the experts about the most important risk/protective fac-tors, and not only with the most stable ranking lists. In terms of predictive power (see Fig.2 in previous sections) RF and Pearson shows similar behavior but the models built with Pearson tend to outperform those built with the features selected with RF. This is also confirmed with the analysis con- ducted in Section 3.2 (Table 2). This visual analysis allows us to see that SVM-wrapper performs moderately in terms of robustness and it is similar to Pearson. Additionally, they are the best ranking techniques in terms of model performance.

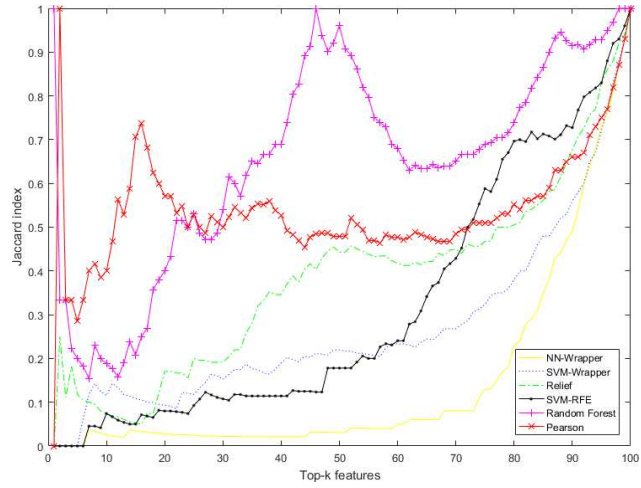

(a)

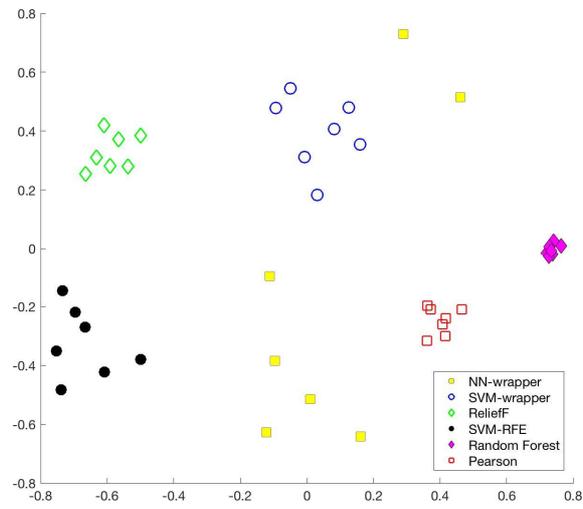

(b)

Figure 7: Feature Selector Stability: (a) Jaccard index for Feature Subsets with different cardinality; (b) MDS plot of the Feature Ranking Algorithms.

## 4. Discussion

### 4.1. Impact

There is enough evidence that different screening tests such as fecal oc- cult blood testing or colonoscopy are effective in reducing the incidence and mortality from CRC [32, 35]. Screening and preventive interventions can benefit from the incorporation of CRC risk prediction models able to iden- tify individuals at high risk of developing CRC. Risk-adapted screening tests might also be more cost-effective than following traditional screening inter- ventions. The use of the individualized risk information provided by these models would also potentially encourage lifestyle changes. There are several challenges, however, to implement risk prediction tools for CRC. The main one is the collection of family history, genetic, lifestyle and dietary informa- tion in a primary care environment. Other side effects include anxiety, false reassurance, and false alarms among the general population. Further assess- ment in terms of research, clinical impact, and cost-effectiveness is necessary to deploy these models in clinical practice.

### 4.2. Contribution

The aim of this study is the assessment of several feature selection tech- niques together with classification models to develop risk prediction models for colorectal cancer. This work is focused in the analysis of both classifica- tion performance and robustness of the feature selection algorithm.

This research work shows that the two best performance results are achieved with a SVM classifier using the top-41 features selected by the SVM-approach (AUC=0.693) and LR with the top-40 features selected by the Pearson (AUC=0.689). This implies an improvement with respect to the results using the full feature set of 3.9% and 1.9% for the SVM and LR classifier, respectively. This performance is comparable to other studies on CRC (AUC=0.63 in [21] and similar performance in references therein).

Table 5: Top-40 Pearson and top-41 SVM-Wrapper feature sets.

| Ranker<br>Classifier<br>AUC | Pearson<br>LR<br>0.689 | SVM-wrapper<br>SVM<br>0.693 | Rank |
|---|---|---|---|
| | Level of education | Family history of CRC | 1 |
| | Age | Level of education | 2 |
| | RED MEAT | Ethanol in the past decade | 3 |
| | Cholesterol | Red Meat | 4 |
| | Ethanol in the past decade | Ethanol in the present | 5 |
| | Polysaccharides | Age | 6 |
| | Family history of CRC | Fiber | 7 |
| | Sex | Physical exercise in the last decade | 8 |
| | Saturated fats | RS 10795668 | 9 |
| | BMI | RS 5934683 | 10 |
| | Total energy | RS 2470890 | 11 |
| | Carbohydrates | RS 10761659 | 12 |
| | Total proteins | RS 2542151 | 13 |
| | Total fats | RS 4444235 | 14 |
| | Animal proteins | RS 4939827 | 15 |
| | Zinc | Legume | 16 |
| | Legume | RS 10936599 | 17 |
| | RS 4939827 | Cholesterol | 18 |
| | Monounsaturated fats | FRUITS | 19 |
| | Niacin | RS 9929218 | 20 |
| | Sodium | RS 5771069 | 21 |
| | Polyunsaturated fats | Sex | 22 |
| | Vegetable proteins | RS 4779584 | 23 |
| | Physical exercise in the last decade | RS 4774302 | 24 |
| | Phosphorus | Vitamin D | 25 |
| | Vegetables | Carotenoids | 26 |
| | Carotenoids | RS 6083 | 27 |
| | Digestible sugars | RS 7014346 | 28 |
| | Edible | RS 961253 | 29 |
| | Thiamin | FISH | 30 |
| | Total grams | RS 1800588 | 31 |
| | Cobalamin | Niacin | 32 |
| | Iron | RS 268 | 33 |
| | RS 961253 | RS 439401 | 34 |
| | RS 9929218 | RS 4925386 | 35 |
| | Retinoids | Race | 36 |
| | RS 7014346 | Total grams | 37 |
| | Cereals | RS 7136702 | 38 |
| | Water | Zinc | 39 |
| | Riboflavin | RS 3802842 | 40 |
| | | RS 9858542 | 41 |

Table 5 shows the features with more discriminant power for colorectal cancer prediction selected with the best performance strategies: the SVM-Wrapper approach (top-41) and the ranking performed with the Pearson correlation coefficient (top-40). Although features are different from one list to the other, some features are common in both lists (highlighted in grey): Red meat, legume, physical exercise, family histoy of CRC, carotenes, cholesterol, age, level of education, ethanol in the past, food intake in grams, niacin, rs4939827, rs7014346, rs961253, rs9929218, sex and zinc. Some of the features found associated with CRC risk in MCC population are known to have a high association with CRC, while others are not considered so correlated, what deserves further study with other cancer data sets.

Stability is assessed in a conventional way by computing an scalar metric. Additionally, we also propose a graphical approach that works in 2D or 3D in order to evaluate not only the stability of the algorithms but also its similarities with other ranking algorithms. This graphical approach based on a MDS projection allows to see at a glance and in a single picture that:
(a) the most stable algorithm is RF, (b) the most unstable is NN-Wrapper, (c) the rankings yielded by RF and Pearson are very similar so that we can focus the analysis on one of them, (d) the before mentioned group leads to a ranking that is very different to Releif and SVM-RFE, (e) the SWM-Wrapper ranking is moderately stable and similar to the Pearson one.

The main strength of this study is that it analyzes stability and predictive power together. Additionally, feature selection techniques allow both the improvement of performance for risk prediction models and the identification of relevant features related to CRC cancer. It turns out that in this study the SVM-wrapper was one of the best ranking technique regarding model performance and it performs moderately in terms of robustness. This study (limited the multicase control-study of the Spanish population) also shows that the simple Pearson correlation coefficient shows a good trade in terms of performance and robustness and can easily scale to high dimensional datasets. A comprehensive evaluation with more colorectal cancer datasets and more feature ranking algorithms can lead to more generalization in this field.

*4.3. Comparison with the state-of-the-art knowledge*

Up to present, there are many attempts aiming at predicting colorectal cancer risk in general population settings [4, 21, 33, 29]. In this section, we assess the performance of a CRC prediction model built with 46 features (29 SNP and 17 environmental) selected by the experts in the field according to the state-of-the-art knowledge [21]. Table 6 shows this list of features.

Table 6: Relevant features according to state-of-the-art knowledge. Features highlighted in bold are those that have also been found relevant in our study.

| | | |
|---|---|---|
| · **Sex** | · **Iron** | · **rs 4774302** |
| · **Age** | · **Fiber** | · **rs 4779584** |
| · **Level of education** | · **Vit D** | · **rs 4925386** |
| | · rs 10411210 | · **rs 4939827** |
| · **Family history of CRC** | · rs 10505477 | · rs 5275 |
| · **Red meat** | · rs 1057910 | · **rs 5934683** |
| · **Vegetables** | · **rs 10795668** | · rs 6687758 |
| · **Ethanol in the past decade** | · **rs 10936599** | · rs 6691170 |
| | · rs 11169552 | · rs 6983267 |
| · **BMI** | · rs 11209026 | · **rs 7014346** |
| · **Physical exercise in the last decade** | · **rs 1800588** | · rs 7136702 |
| | · rs 1801282 | · rs 744166 |
| · **Legume** | · rs 3802842 | · rs 762551 |
| · **Fruit** | · rs 3824999 | · rs 7758229 |
| · **Fish** | · rs 405509 | · **rs 961253** |
| · Dairy | · **rs 4444235** | · **rs 9929218** |
| · **Energy** | | |

Each one of the classifiers considered in this study was assessed in terms of AUC with four different feature sets (see Table 7): The set selected by the experts (46 features), a feature set with the 28 variables that are common to the Experts's set and the Top-40 union (that is, keeping only the features suggested by the experts that were found in a relevant position in our study) and finally, the Top-40 union set (64 features that appear in Table 5).

Table 7: AUC for different classifiers with different feature sets.

| Feature Set | Cardinality | LR | k-NN | NN | SVM | BT |
|---|---|---|---|---|---|---|
| Full feature set | 100 | 0.676 | 0.624 | 0.540 | 0.667 | 0.671 |
| Experts' set | 46 | 0.679 | 0.636 | 0.545 | 0.652 | 0.661 |
| Experts' set ∩ Top-40 Union | 28 | 0.683 | 0.653 | 0.561 | 0.667 | 0.660 |
| Top-40 Union | 64 | 0.686 | 0.653 | 0.583 | 0.686 | 0.676 |

It can be observed that the performance is completely unaffected when the feature set is reduced from 46 features provided by the experts to 28 fea- tures. Removing the features that were not found relevant in our study either maintains or increases the AUC (Table 7). Thus, AUC for the SVM classifier increases from 0.652 to 0.667 and from 0.679 to 0.683 for the LR approach. It is noteworthy that performance never decreases in spite of removing features that were considered relevant in the literature. If we increase the feature set to a cardinality of 64 with the next most relevant features (Top-40 union), performance shows the same behavior, that is, either increasing or maintain- ing its value. Thus, for example, it increases from 0.667 with 28 features to 0.686 with the Top-40 union for the SVM classifier, compared with an AUC of 0.652 achieved with the Experts' set. Likewise, the AUC for the BT clas- sifier is 0.676 compared with 0.660 when using less features or the experts' feature set. The Top-40 union set leads to an increase in performance with respect to the full feature set and the experts' feature subset and on average to the best performance results.

This analysis suggests that some of the features proposed by the experts are either redundant or irrelevant since performance is not affected by re- moving them. This also is confirmed by the fact that these features do not hold top positions in the ranking lists obtained in our experimental setting. This is the case of the dairy consumption (it could be redundant since VitD is also in the list) or SNPs such as, rs6983267, rs10411210 or rs7758229.

Note that the AUC prediction results achieved with the experts' feature set compared with the full set leads to an increase in performance of 0.4% for LR and 1.9% for k-NN. This improvement, though, is less than the achieved with feature selection strategies, which is 1.9% for LR and 7.8% for k-NN. Performance with the expert set, however, drops $-2.2\%$ and $-1.5\%$ for the SVM and BT models, while it is always increased with feature selection algorithms. This could indicate that some features excluded from the list should be given more relevance in this context.

When comparing our results with the features provided by the experts according to the state-of-the-art knowledge, it turns out that both lists have in common many features.The features highlighted in bold in Table 6 are those that also appear in any of the selected top-41 SVM-wrapper and top-40 Pearson lists . Note that almost two out of three variables are also in our reduced set of the most relevant features.

Other features, though, suggested by the experts (some SNPs) do not seem to affect the predictive power of the model and some features like Zinc, Carotenoids, Niacin that were found relevant in our experimental setting are not considered relevant in the literature, which deserves further study.

*4.4. Limitations*

We acknowledge that this study is focused on the Spanish population and our findings may not directly translate to individuals with other ethnic- ities. The study included six feature ranking algorithms selected from the three main categories (filter, wrapper and embedded), but there are many more different widely used feature selection algorithms that could be tested. However, results may still be widely applicable as they reconfirm previous findings and point out new factors to be considered in further studies.

*4.5. Future work*

Future work includes the study of ensemble strategies to increase the stability of feature selection techniques, in particular those that have a high margin of improvement. Since ranking algorithms may have a high computational cost, our aim is also to explore new hybrid ranking approaches based on two steps: (1) a first simple one based on filters able to quickly remove the most irrelevant features and (2) a second phase with wrapper or embedded ranking algorithms focused on the subset of features selected in the first step. We consider to test these techniques on a global dataset with thousands of SNPS and more instances. Having access to a bigger dataset, we also aim to assess deep learning approaches that have shown outstanding performance in many filed.

## 5. Conclusions

Appropriate feature selection is required when building a colorectal cancer risk prediction model. It helps to avoid overfitting and is an aid to identify the features with more prediction power so that proper interventions can be taken to address the risk. Assessing the stability of the feature selection methods becomes necessary, otherwise conclusions derived from the analysis may be quite unreliable. The graphical approach that is presented here enables us to analyze the stability of feature selection algorithms as well as the similarity among different feature ranking techniques.

Comparisons have been conducted with several feature ranking algorithms and different risk prediction models. The experimental results on the multi-case control-study of the Spanish population indicate that the SVM-wrapper approach shows moderate stability and it leads to the best classification model performance. In addition, the simple Pearson correlation coefficient shows a good trade in terms of performance and stability.

Screening and preventive interventions can certainly benefit from an improved estimation of the risk of developing CRC. However, there are still some barriers and more research to be done in order to incorporate it into a daily clinical practice.

**Disclosure statements**

MCC-Spain Study Group: G. Castaño-Vinyals, B. Pérez-Gómez, J. Llorca, J. M. Altzibar, E. Ardanaz, S. de Sanjosé, J.J. Jiménez-Moleón, A. Tardón, J. Alguacil, R. Peiró, R. Marcos-Gragera, C. Navarro, M. Pollán and M. Kogevinas.

**Acknowledgements**

All the subjects who participated in the study and all MCC-Spain colaborators.

**Funding**

The study was partially funded by the Accion Transversal del Cancer, approved on the Spanish Ministry Council on the 11th October 2007, by the Instituto de Salud Carlos III-FEDER (PI08/1770, PI08/0533, PI08/1359, PS09/00773, PS09/01286, PS09/01903, PS09/02078, PS09/01662, PI11/01403, PI11/01889, PI11/00226, PI11/01810, PI11/02213, PI12/00488, PI12/00265, PI12/01270, PI12/00715, PI12/00150), by the Fundacin Marqus de Valdecilla (API 10/09), by the ICGC International Cancer Genome Consortium CLL, by the Junta de Castilla y León (LE22A10-2), by the Consejera de Salud of the Junta de Andalucía (PI-0571), by the Conselleria de Sanitat of the Generalitat Valenciana (AP 061/10), by the Recercaixa (2010ACUP 00310), by the Regional Government of the Basque Country by European Commission grants FOOD-CT- 2006-036224- HI- WATE, by the Spanish Association Against Cancer (AECC) Scientific Foundation, by the The Catalan Government DURSI grant 2009SGR1489. Samples: Biological samples were stored at the Parc de Salut MAR Biobank (MARBiobanc; Barcelona) which is supported by Instituto de Salud Carlos III FEDER (RD09/0076/00036). Also at the Public Health Laboratory from Gipuzkoa and the Basque Biobank. Also sample collection was supported by the Xarxa de Bancs de Tumors de Catalunya sponsored by Pla Director dOncologia de Catalunya (XBTC). Biological samples were stored at the Biobanco La Fe which is supported by Instituto de Salud Carlos III (RD 09 0076/00021) and FISABIO biobanking, which is supported by Instituto de Salud Carlos III (RD09 0076/00058).

**Genotyping**: SNP genotyping services were provided by the Spanish Centro Nacional de Genotipado (CEGEN-ISCIII)" and by the Basque Biobank.

## Conflict of interest

The authors declare that they have no conflicts of interest.